\documentclass[letterpaper, 10 pt, conference]{ieeeconf}
\IEEEoverridecommandlockouts
\usepackage{cite}
\usepackage{amsmath,amssymb,amsfonts}
\usepackage{graphicx}
\usepackage{textcomp}
\usepackage{xcolor}
\usepackage{comment}
\def\BibTeX{{\rm B\kern-.05em{\sc i\kern-.025em b}\kern-.08em
    T\kern-.1667em\lower.7ex\hbox{E}\kern-.125emX}}

\usepackage{subcaption}
\usepackage{ragged2e}
\usepackage[noend]{algpseudocode}
\usepackage{algorithm}%

\usepackage{amssymb}
\usepackage{amsfonts}
\usepackage{mathrsfs}
\usepackage{xspace}
\usepackage{bm}
\usepackage{upgreek}

\newcommand{\safemath}[2]{\newcommand{#1}{\ensuremath{#2}\xspace}}

\safemath{\bma}{\mathbf{a}}
\safemath{\bmb}{\mathbf{b}}
\safemath{\bmc}{\mathbf{c}}
\safemath{\bmd}{\mathbf{d}}
\safemath{\bme}{\mathbf{e}}
\safemath{\bmf}{\mathbf{f}}
\safemath{\bmg}{\mathbf{g}}
\safemath{\bmh}{\mathbf{h}}
\safemath{\bmi}{\mathbf{i}}
\safemath{\bmj}{\mathbf{j}}
\safemath{\bmk}{\mathbf{k}}
\safemath{\bml}{\mathbf{l}}
\safemath{\bmm}{\mathbf{m}}
\safemath{\bmn}{\mathbf{n}}
\safemath{\bmo}{\mathbf{o}}
\safemath{\bmp}{\mathbf{p}}
\safemath{\bmq}{\mathbf{q}}
\safemath{\bmr}{\mathbf{r}}
\safemath{\bms}{\mathbf{s}}
\safemath{\bmt}{\mathbf{t}}
\safemath{\bmu}{\mathbf{u}}
\safemath{\bmv}{\mathbf{v}}
\safemath{\bmw}{\mathbf{w}}
\safemath{\bmx}{\mathbf{x}}
\safemath{\bmy}{\mathbf{y}}
\safemath{\bmz}{\mathbf{z}}
\safemath{\bmzero}{\mathbf{0}}
\safemath{\bmone}{\mathbf{1}}

\bmdefine{\biad}{a}
\bmdefine{\bibd}{b}
\bmdefine{\bicd}{c}
\bmdefine{\bidd}{d}
\bmdefine{\bied}{e}
\bmdefine{\bifd}{f}
\bmdefine{\bigd}{g}
\bmdefine{\bihd}{h}
\bmdefine{\biid}{i}
\bmdefine{\bijd}{j}
\bmdefine{\bikd}{k}
\bmdefine{\bild}{l}
\bmdefine{\bimd}{m}
\bmdefine{\bind}{n}
\bmdefine{\biod}{o}
\bmdefine{\bipd}{p}
\bmdefine{\biqd}{q}
\bmdefine{\bird}{r}
\bmdefine{\bisd}{s}
\bmdefine{\bitd}{t}
\bmdefine{\biud}{u}
\bmdefine{\bivd}{v}
\bmdefine{\biwd}{w}
\bmdefine{\bixd}{x}
\bmdefine{\biyd}{y}
\bmdefine{\bizd}{z}

\bmdefine{\bixid}{\xi}
\bmdefine{\bilambdad}{\lambda}
\bmdefine{\bimud}{\mu}
\bmdefine{\bithetad}{\theta}
\bmdefine{\biphid}{\phi}
\bmdefine{\bideltad}{\delta}

\safemath{\bmia}{\biad}
\safemath{\bmib}{\bibd}
\safemath{\bmic}{\bicd}
\safemath{\bmid}{\bidd}
\safemath{\bmie}{\bied}
\safemath{\bmif}{\bifd}
\safemath{\bmig}{\bigd}
\safemath{\bmih}{\bihd}
\safemath{\bmii}{\biid}
\safemath{\bmij}{\bijd}
\safemath{\bmik}{\bikd}
\safemath{\bmil}{\bild}
\safemath{\bmim}{\bimd}
\safemath{\bmin}{\bind}
\safemath{\bmio}{\biod}
\safemath{\bmip}{\bipd}
\safemath{\bmiq}{\biqd}
\safemath{\bmir}{\bird}
\safemath{\bmis}{\bisd}
\safemath{\bmit}{\bitd}
\safemath{\bmiu}{\biud}
\safemath{\bmiv}{\bivd}
\safemath{\bmiw}{\biwd}
\safemath{\bmix}{\bixd}
\safemath{\bmiy}{\biyd}
\safemath{\bmiz}{\bizd}

\safemath{\bmxi}{\bixid}
\safemath{\bmlambda}{\bilambdad}
\safemath{\bmmu}{\bimud}
\safemath{\bmtheta}{\bithetad}
\safemath{\bmphi}{\biphid}
\safemath{\bmdelta}{\bideltad}

\safemath{\bA}{\mathbf{A}}
\safemath{\bB}{\mathbf{B}}
\safemath{\bC}{\mathbf{C}}
\safemath{\bD}{\mathbf{D}}
\safemath{\bE}{\mathbf{E}}
\safemath{\bF}{\mathbf{F}}
\safemath{\bG}{\mathbf{G}}
\safemath{\bH}{\mathbf{H}}
\safemath{\bI}{\mathbf{I}}
\safemath{\bJ}{\mathbf{J}}
\safemath{\bK}{\mathbf{K}}
\safemath{\bL}{\mathbf{L}}
\safemath{\bM}{\mathbf{M}}
\safemath{\bN}{\mathbf{N}}
\safemath{\bO}{\mathbf{O}}
\safemath{\bP}{\mathbf{P}}
\safemath{\bQ}{\mathbf{Q}}
\safemath{\bR}{\mathbf{R}}
\safemath{\bS}{\mathbf{S}}
\safemath{\bT}{\mathbf{T}}
\safemath{\bU}{\mathbf{U}}
\safemath{\bV}{\mathbf{V}}
\safemath{\bW}{\mathbf{W}}
\safemath{\bX}{\mathbf{X}}
\safemath{\bY}{\mathbf{Y}}
\safemath{\bZ}{\mathbf{Z}}

\safemath{\bZero}{\mathbf{0}}
\safemath{\bOne}{\mathbf{1}}
\safemath{\bDelta}{\mathbf{\Delta}}
\safemath{\bLambda}{\boldsymbol\Lambda}
\safemath{\bPhi}{\mathbf{\Upphi}}
\safemath{\bSigma}{\mathbf{\Upsigma}}
\safemath{\bOmega}{\mathbf{\Upomega}}
\safemath{\bTheta}{\mathbf{\Uptheta}}

\bmdefine{\biAd}{A}
\bmdefine{\biBd}{B}
\bmdefine{\biCd}{C}
\bmdefine{\biDd}{D}
\bmdefine{\biEd}{E}
\bmdefine{\biFd}{F}
\bmdefine{\biGd}{G}
\bmdefine{\biHd}{H}
\bmdefine{\biId}{I}
\bmdefine{\biJd}{J}
\bmdefine{\biKd}{K}
\bmdefine{\biLd}{L}
\bmdefine{\biMd}{M}
\bmdefine{\biOd}{N}
\bmdefine{\biPd}{O}
\bmdefine{\biQd}{P}
\bmdefine{\biRd}{R}
\bmdefine{\biSd}{S}
\bmdefine{\biTd}{T}
\bmdefine{\biUd}{U}
\bmdefine{\biVd}{V}
\bmdefine{\biWd}{W}
\bmdefine{\biXd}{X}
\bmdefine{\biYd}{Y}
\bmdefine{\biZd}{Z}

\bmdefine{\biDelta}{\Delta}
\bmdefine{\biLambda}{\Lambda}
\bmdefine{\biPhi}{\Phi}
\bmdefine{\biSigma}{\Sigma}
\bmdefine{\biOmega}{\Omega}
\bmdefine{\biTheta}{\Theta}

\safemath{\bimA}{\biAd}
\safemath{\bimB}{\biBd}
\safemath{\bimC}{\biCd}
\safemath{\bimD}{\biDd}
\safemath{\bimE}{\biEd}
\safemath{\bimF}{\biFd}
\safemath{\bimG}{\biGd}
\safemath{\bimH}{\biHd}
\safemath{\bimI}{\biId}
\safemath{\bimJ}{\biJd}
\safemath{\bimK}{\biKd}
\safemath{\bimL}{\biLd}
\safemath{\bimM}{\biMd}
\safemath{\bimN}{\biNd}
\safemath{\bimO}{\biOd}
\safemath{\bimP}{\biPd}
\safemath{\bimQ}{\biQd}
\safemath{\bimR}{\biRd}
\safemath{\bimS}{\biSd}
\safemath{\bimT}{\biTd}
\safemath{\bimU}{\biUd}
\safemath{\bimV}{\biVd}
\safemath{\bimW}{\biWd}
\safemath{\bimX}{\biXd}
\safemath{\bimY}{\biYd}
\safemath{\bimZ}{\biZd}

\safemath{\bimDelta}{\biDelta}
\safemath{\bimLambda}{\biLambda}
\safemath{\bimPhi}{\biPhi}
\safemath{\bimSigma}{\biSigma}
\safemath{\bimOmega}{\biOmega}
\safemath{\bimTheta}{\biTheta}

\safemath{\setA}{\mathcal{A}}
\safemath{\setB}{\mathcal{B}}
\safemath{\setC}{\mathcal{C}}
\safemath{\setD}{\mathcal{D}}
\safemath{\setE}{\mathcal{E}}
\safemath{\setF}{\mathcal{F}}
\safemath{\setG}{\mathcal{G}}
\safemath{\setH}{\mathcal{H}}
\safemath{\setI}{\mathcal{I}}
\safemath{\setJ}{\mathcal{J}}
\safemath{\setK}{\mathcal{K}}
\safemath{\setL}{\mathcal{L}}
\safemath{\setM}{\mathcal{M}}
\safemath{\setN}{\mathcal{N}}
\safemath{\setO}{\mathcal{O}}
\safemath{\setP}{\mathcal{P}}
\safemath{\setQ}{\mathcal{Q}}
\safemath{\setR}{\mathcal{R}}
\safemath{\setS}{\mathcal{S}}
\safemath{\setT}{\mathcal{T}}
\safemath{\setU}{\mathcal{U}}
\safemath{\setV}{\mathcal{V}}
\safemath{\setW}{\mathcal{W}}
\safemath{\setX}{\mathcal{X}}
\safemath{\setY}{\mathcal{Y}}
\safemath{\setZ}{\mathcal{Z}}
\safemath{\emptySet}{\varnothing}

\safemath{\colA}{\mathscr{A}}
\safemath{\colB}{\mathscr{B}}
\safemath{\colC}{\mathscr{C}}
\safemath{\colD}{\mathscr{D}}
\safemath{\colE}{\mathscr{E}}
\safemath{\colF}{\mathscr{F}}
\safemath{\colG}{\mathscr{G}}
\safemath{\colH}{\mathscr{H}}
\safemath{\colI}{\mathscr{I}}
\safemath{\colJ}{\mathscr{J}}
\safemath{\colK}{\mathscr{K}}
\safemath{\colL}{\mathscr{L}}
\safemath{\colM}{\mathscr{M}}
\safemath{\colN}{\mathscr{N}}
\safemath{\colO}{\mathscr{O}}
\safemath{\colP}{\mathscr{P}}
\safemath{\colQ}{\mathscr{Q}}
\safemath{\colR}{\mathscr{R}}
\safemath{\colS}{\mathscr{S}}
\safemath{\colT}{\mathscr{T}}
\safemath{\colU}{\mathscr{U}}
\safemath{\colV}{\mathscr{V}}
\safemath{\colW}{\mathscr{W}}
\safemath{\colX}{\mathscr{X}}
\safemath{\colY}{\mathscr{Y}}
\safemath{\colZ}{\mathscr{Z}}

\safemath{\opA}{\mathbb{A}}
\safemath{\opB}{\mathbb{B}}
\safemath{\opC}{\mathbb{C}}
\safemath{\opD}{\mathbb{D}}
\safemath{\opE}{\mathbb{E}}
\safemath{\opF}{\mathbb{F}}
\safemath{\opG}{\mathbb{G}}
\safemath{\opH}{\mathbb{H}}
\safemath{\opI}{\mathbb{I}}
\safemath{\opJ}{\mathbb{J}}
\safemath{\opK}{\mathbb{K}}
\safemath{\opL}{\mathbb{L}}
\safemath{\opM}{\mathbb{M}}
\safemath{\opN}{\mathbb{N}}
\safemath{\opO}{\mathbb{O}}
\safemath{\opP}{\mathbb{P}}
\safemath{\opQ}{\mathbb{Q}}
\safemath{\opR}{\mathbb{R}}
\safemath{\opS}{\mathbb{S}}
\safemath{\opT}{\mathbb{T}}
\safemath{\opU}{\mathbb{U}}
\safemath{\opV}{\mathbb{V}}
\safemath{\opW}{\mathbb{W}}
\safemath{\opX}{\mathbb{X}}
\safemath{\opY}{\mathbb{Y}}
\safemath{\opZ}{\mathbb{Z}}
\safemath{\opZero}{\mathbb{O}}
\safemath{\identityop}{\opI}

\safemath{\veca}{\bma}
\safemath{\vecb}{\bmb}
\safemath{\vecc}{\bmc}
\safemath{\vecd}{\bmd}
\safemath{\vece}{\bme}
\safemath{\vecf}{\bmf}
\safemath{\vecg}{\bmg}
\safemath{\vech}{\bmh}
\safemath{\veci}{\bmi}
\safemath{\vecj}{\bmj}
\safemath{\veck}{\bmk}
\safemath{\vecl}{\bml}
\safemath{\vecm}{\bmm}
\safemath{\vecn}{\bmn}
\safemath{\veco}{\bmo}
\safemath{\vecp}{\bmp}
\safemath{\vecq}{\bmq}
\safemath{\vecr}{\bmr}
\safemath{\vecs}{\bms}
\safemath{\vect}{\bmt}
\safemath{\vecu}{\bmu}
\safemath{\vecv}{\bmv}
\safemath{\vecw}{\bmw}
\safemath{\vecx}{\bmx}
\safemath{\vecy}{\bmy}
\safemath{\vecz}{\bmz}

\safemath{\veczero}{\bmzero}
\safemath{\vecone}{\bmone}
\safemath{\vecxi}{\bmxi}
\safemath{\veclambda}{\bmlambda}
\safemath{\vecmu}{\bmmu}
\safemath{\vectheta}{\bmtheta}
\safemath{\vecphi}{\bmphi}
\safemath{\vecdelta}{\bmdelta}

\safemath{\matA}{\bA}
\safemath{\matB}{\bB}
\safemath{\matC}{\bC}
\safemath{\matD}{\bD}
\safemath{\matE}{\bE}
\safemath{\matF}{\bF}
\safemath{\matG}{\bG}
\safemath{\matH}{\bH}
\safemath{\matI}{\bI}
\safemath{\matJ}{\bJ}
\safemath{\matK}{\bK}
\safemath{\matL}{\bL}
\safemath{\matM}{\bM}
\safemath{\matN}{\bN}
\safemath{\matO}{\bO}
\safemath{\matP}{\bP}
\safemath{\matQ}{\bQ}
\safemath{\matR}{\bR}
\safemath{\matS}{\bS}
\safemath{\matT}{\bT}
\safemath{\matU}{\bU}
\safemath{\matV}{\bV}
\safemath{\matW}{\bW}
\safemath{\matX}{\bX}
\safemath{\matY}{\bY}
\safemath{\matZ}{\bZ}
\safemath{\matzero}{\bmzero}

\safemath{\matDelta}{\bDelta}
\safemath{\matLambda}{\bLambda}
\safemath{\matPhi}{\bPhi}
\safemath{\matSigma}{\bSigma}
\safemath{\matOmega}{\bOmega}
\safemath{\matTheta}{\bTheta}

\safemath{\matidentity}{\matI}
\safemath{\matone}{\matO}

\safemath{\rnda}{A}
\safemath{\rndb}{B}
\safemath{\rndc}{C}
\safemath{\rndd}{D}
\safemath{\rnde}{E}
\safemath{\rndf}{F}
\safemath{\rndg}{G}
\safemath{\rndh}{H}
\safemath{\rndi}{I}
\safemath{\rndj}{J}
\safemath{\rndk}{K}
\safemath{\rndl}{L}
\safemath{\rndm}{M}
\safemath{\rndn}{N}
\safemath{\rndo}{O}
\safemath{\rndp}{P}
\safemath{\rndq}{Q}
\safemath{\rndr}{R}
\safemath{\rnds}{S}
\safemath{\rndt}{T}
\safemath{\rndu}{U}
\safemath{\rndv}{V}
\safemath{\rndw}{W}
\safemath{\rndx}{X}
\safemath{\rndy}{Y}
\safemath{\rndz}{Z}

\safemath{\rveca}{\bimA}
\safemath{\rvecb}{\bimB}
\safemath{\rvecc}{\bimC}
\safemath{\rvecd}{\bimD}
\safemath{\rvece}{\bimE}
\safemath{\rvecf}{\bimF}
\safemath{\rvecg}{\bimG}
\safemath{\rvech}{\bimH}
\safemath{\rveci}{\bimI}
\safemath{\rvecj}{\bimJ}
\safemath{\rveck}{\bimK}
\safemath{\rvecl}{\bimL}
\safemath{\rvecm}{\bimM}
\safemath{\rvecn}{\bimN}
\safemath{\rveco}{\bomO}
\safemath{\rvecp}{\bimP}
\safemath{\rvecq}{\bimQ}
\safemath{\rvecr}{\bimR}
\safemath{\rvecs}{\bimS}
\safemath{\rvect}{\bimT}
\safemath{\rvecu}{\bimU}
\safemath{\rvecv}{\bimV}
\safemath{\rvecw}{\bimW}
\safemath{\rvecx}{\bimX}
\safemath{\rvecy}{\bimY}
\safemath{\rvecz}{\bimZ}

\safemath{\rvecxi}{\bmxi}
\safemath{\rveclambda}{\bmlambda}
\safemath{\rvecmu}{\bmmu}
\safemath{\rvectheta}{\bmtheta}
\safemath{\rvecphi}{\bmphi}

\safemath{\rmatA}{\bimA}
\safemath{\rmatB}{\bimB}
\safemath{\rmatC}{\bimC}
\safemath{\rmatD}{\bimD}
\safemath{\rmatE}{\bimE}
\safemath{\rmatF}{\bimF}
\safemath{\rmatG}{\bimG}
\safemath{\rmatH}{\bimH}
\safemath{\rmatI}{\bimI}
\safemath{\rmatJ}{\bimJ}
\safemath{\rmatK}{\bimK}
\safemath{\rmatL}{\bimL}
\safemath{\rmatM}{\bimM}
\safemath{\rmatN}{\bimN}
\safemath{\rmatO}{\bimO}
\safemath{\rmatP}{\bimP}
\safemath{\rmatQ}{\bimQ}
\safemath{\rmatR}{\bimR}
\safemath{\rmatS}{\bimS}
\safemath{\rmatT}{\bimT}
\safemath{\rmatU}{\bimU}
\safemath{\rmatV}{\bimV}
\safemath{\rmatW}{\bimW}
\safemath{\rmatX}{\bimX}
\safemath{\rmatY}{\bimY}
\safemath{\rmatZ}{\bimZ}

\safemath{\rmatDelta}{\bimDelta}
\safemath{\rmatLambda}{\bimLambda}
\safemath{\rmatPhi}{\bimPhi}
\safemath{\rmatSigma}{\bimSigma}
\safemath{\rmatOmega}{\bimOmega}
\safemath{\rmatTheta}{\bimTheta}

\usepackage{amssymb}
\usepackage{amsfonts}
\usepackage{mathrsfs}
\usepackage{xspace}
\usepackage{bm}
\usepackage{fancyref}
\usepackage{textcomp}

\usepackage{multirow}
\usepackage{stmaryrd}

\newenvironment{textbmatrix}{	\setlength{\arraycolsep}{2.5pt}%
								\big[\begin{matrix}}{\end{matrix}\big]%
								\raisebox{0.08ex}{\vphantom{M}}}

\def\be{\begin{equation}}
\def\ee{\end{equation}}
\def\een{\nonumber \end{equation}}
\def\mat{\begin{bmatrix}}
\def\emat{\end{bmatrix}}
\def\btm{\begin{textbmatrix}}
\def\etm{\end{textbmatrix}}

\def\ba#1\ea{\begin{align}#1\end{align}}
\def\bas#1\eas{\begin{align*}#1\end{align*}}
\def\bs#1\es{\begin{split}#1\end{split}} 
\def\bg#1\eg{\begin{gather}#1\end{gather}}
\def\bml#1\eml{\begin{multline}#1\end{multline}}
\def\bi#1\ei{\begin{itemize}#1\end{itemize}}

\DeclareMathOperator*{\argmin}{arg\;min}		%

\safemath{\dirac}{\delta}					%
\safemath{\krond}{\dirac}					%
\safemath{\upto}{\uparrow}
\safemath{\downto}{\downarrow}
\safemath{\iu}{j}							%
\safemath{\ev}{\lambda}						%
\safemath{\hilseqspace}{l^{2}}				%
\newcommand{\banachfunspace}[1]{\setL^{#1}}	%
\safemath{\hilfunspace}{\banachfunspace{2}}	%

\safemath{\SNR}{\textsf{SNR}} 				%
\safemath{\PAR}{\textsf{PAR}} 				%
\safemath{\No}{N_0}							%
\safemath{\Es}{E_s}							%
\safemath{\Eb}{E_b}							%
\safemath{\EbNo}{\frac{\Eb}{\No}}
\safemath{\EsNo}{\frac{\Es}{\No}}

\DeclareMathOperator{\CHop}{\ensuremath{\opH}} %
\safemath{\tvir}{\rndh_{\CHop}}				%
\safemath{\tvtf}{\rndl_{\CHop}}				%
\safemath{\spf}{\rnds_{\CHop}}				%
\safemath{\bff}{H_{\CHop}}					%

\safemath{\ircf}{r_{h}}						%
\safemath{\tftvcf}{r_{s}}					%
\safemath{\tfcf}{r_{l}}						%
\safemath{\bfcf}{r_{H}}						%

\safemath{\tcorr}{c_h}						%
\safemath{\scf}{c_{s}}						%
\safemath{\tfcorr}{c_{l}}					%
\safemath{\fcorr}{c_{H}}						%

\safemath{\mi}{I}							%
\safemath{\capacity}{C}						%

\safemath{\normal}{\mathcal{N}}			%
\safemath{\jpg}{\mathcal{CN}}			%
\safemath{\mchain}{\leftrightarrow}		%

\safemath{\dB}{\,\mathrm{dB}}
\safemath{\dBm}{\,\mathrm{dBm}}
\safemath{\Hz}{\,\mathrm{Hz}}
\safemath{\kHz}{\,\mathrm{kHz}}
\safemath{\MHz}{\,\mathrm{MHz}}
\safemath{\GHz}{\,\mathrm{GHz}}
\safemath{\s}{\,\mathrm{s}}
\safemath{\ms}{\,\mathrm{ms}}
\safemath{\mus}{\,\mathrm{\text{\textmu}s}}
\safemath{\ns}{\,\mathrm{ns}}
\safemath{\ps}{\,\mathrm{ps}}
\safemath{\meter}{\,\mathrm{m}}
\safemath{\mm}{\,\mathrm{mm}}
\safemath{\cm}{\,\mathrm{cm}}
\safemath{\m}{\,\mathrm{m}}
\safemath{\W}{\,\mathrm{W}}
\safemath{\mW}{\, \mathrm{mW}}
\safemath{\J}{\,\mathrm{J}}
\safemath{\K}{\,\mathrm{K}}
\safemath{\bit}{\,\mathrm{bit}}
\safemath{\nat}{\,\mathrm{nat}}

\safemath{\define}{\triangleq}			%

\safemath{\equivalent}{\sim}
\safemath{\distas}{\sim}					%
\safemath{\sdiff}{\Delta}				%

\safemath{\reals}{\mathbb{R}}
\safemath{\positivereals}{\reals_{+}}
\safemath{\integers}{\mathbb{Z}}
\safemath{\posint}{\integers_{+}}
\safemath{\naturals}{\mathbb{N}}
\safemath{\posnaturals}{\naturals_{+}}
\safemath{\complexset}{\mathbb{C}}
\safemath{\rationals}{\mathbb{Q}}

\newcommand*{\fancyrefapplabelprefix}{app}		%
\newcommand*{\fancyrefthmlabelprefix}{thm}		%
\newcommand*{\fancyreflemlabelprefix}{lem}		%
\newcommand*{\fancyrefcorlabelprefix}{cor}		%
\newcommand*{\fancyrefdeflabelprefix}{def}		%
\newcommand*{\fancyrefproplabelprefix}{prop}	%
\newcommand*{\fancyrefobslabelprefix}{obs}		%
\newcommand*{\fancyrefalglabelprefix}{alg}		%
\newcommand*{\fancyrefasmlabelprefix}{asm}	    %
\newcommand*{\fancyrefasmslabelprefix}{asms}	    %
\newcommand*{\fancyreftbllabelprefix}{tbl}	    %
\newcommand*{\fancyrefestilabelprefix}{esti}	    %

\frefformat{vario}{\fancyrefseclabelprefix}{Section~#1}
\frefformat{vario}{\fancyrefthmlabelprefix}{Theorem~#1}
\frefformat{vario}{\fancyreflemlabelprefix}{Lemma~#1}
\frefformat{vario}{\fancyrefcorlabelprefix}{Corollary~#1}
\frefformat{vario}{\fancyrefdeflabelprefix}{Definition~#1}
\frefformat{vario}{\fancyrefobslabelprefix}{Observation~#1}
\frefformat{vario}{\fancyrefasmlabelprefix}{Assumption~#1}
\frefformat{vario}{\fancyrefasmslabelprefix}{Assumptions~#1}
\frefformat{vario}{\fancyreffiglabelprefix}{Figure~#1}
\frefformat{vario}{\fancyrefapplabelprefix}{Appendix~#1} 
\frefformat{vario}{\fancyrefproplabelprefix}{Proposition~#1}
\frefformat{vario}{\fancyrefalglabelprefix}{Algorithm~#1}
\frefformat{vario}{\fancyrefeqlabelprefix}{(#1)}
\frefformat{vario}{\fancyreftbllabelprefix}{Table~#1}
\frefformat{vario}{\fancyrefestilabelprefix}{Estimator~#1}

\newcommand{\algoname}{\mathrm{DecSTER}}%

\newcommand{\targetstate}{\textbf{x}}

\newcommand{\targetposx}{l_x}%
\newcommand{\targetposy}{l_y}%
\newcommand{\targetvelx}{v_x}%
\newcommand{\targetvely}{v_y}%
\newcommand{\nlength}{n_l}%
\newcommand{\nwidth}{n_w}%
\newcommand{\maxspeed}{v_{\text{max}}}
\newcommand{\teamsize}{J}
\newcommand{\iagent}{j}
\newcommand{\dataset}{\mathbb{D}}
\newcommand{\itime}{t}
\newcommand{\measurestate}[1]{h(#1)}
\newcommand{\measurenoise}{\omega}
\newcommand{\numtargets}{k}
\newcommand{\searchspace}{\mathcal{G}}
\newcommand{\totaltime}{T}
\newcommand{\action}{\bma}

\newcommand{\transpose}{\text{T}}

\newcommand{\expectation}{\mathbb{E}}

\safemath{\dictab}{[\,\dicta\,\,\dictb\,]}

\safemath{\ysig}{\bmy}
\safemath{\ysighat}{\hat{\ysig}}
\safemath{\ysigdim}{M}
\safemath{\xsig}{\bmx}
\safemath{\xsigdim}{N}
\safemath{\nx}{n_x}
\safemath{\zsig}{\bmz}
\safemath{\zsigdim}{\ysigdim}
\safemath{\rsig}{\bmr}
\safemath{\Adict}{\bA}
\safemath{\Adicttilde}{\widetilde{\Adict}}
\safemath{\Adictdim}{\outputdim\times\xsigdim}
\safemath{\avec}{\bma}
\safemath{\avectilde}{\tilde{\avec}}
\safemath{\Bdict}{\bB}
\safemath{\Bdicttilde}{\widetilde{\Bdict}}
\safemath{\Cdict}{\bC}
\safemath{\cvec}{\bmc}
\safemath{\Ddict}{\bD}
\safemath{\Ddictdim}{\ysigdim\times\xsigdim}
\safemath{\dvec}{\bmd}
\safemath{\Ddicttilde}{\widetilde{\bD}}
\safemath{\Bonb}{\bB}
\safemath{\bvec}{\bmb}
\safemath{\Bonbdim}{\ysigdim\times\ysigdim}
\safemath{\noise}{\bmn}
\safemath{\noisedim}{\ysigim}
\safemath{\err}{\bme}
\safemath{\errdim}{\ysigdim}
\safemath{\errset}{\setE}
\safemath{\nerr}{n_e}
\safemath{\delop}{\bP_\errset}
\safemath{\delopc}{\bP_{{\errset}^c}}

\safemath{\cplxi}{\imath}
\safemath{\cplxj}{\jmath}

\safemath{\dict}{\matD}
\safemath{\inputdim}{N}		%
\safemath{\outputdim}{M}		%
\safemath{\sparsity}{S}	%
\safemath{\inputdimA}{{N_a}}	%
\safemath{\inputdimB}{{N_b}}	%
\safemath{\elemA}{{n_a}}	%
\safemath{\elemB}{{n_b}}	%
\safemath{\resA}{\matR_a}	%
\safemath{\resB}{\matR_b}	%
\safemath{\subD}{\matS} %
\safemath{\subA}{\matS_a} %
\safemath{\subB}{\matS_b} %
\safemath{\dicta}{\matA} 	%
\safemath{\dictb}{\matB} 	%
\safemath{\hollowS}{H}
\safemath{\hollowA}{H_a}
\safemath{\hollowB}{H_b}
\safemath{\cross}{Z}
\safemath{\coh}{\mu_d}			%
\safemath{\coha}{\mu_a}			%
\safemath{\cohb}{\mu_b}			%
\safemath{\mubs}{\nu}	%
\safemath{\cohm}{\mu_m} %
\safemath{\dictset}{\setD}	%
\safemath{\dictsetp}{\dictset(\coh,\coha,\cohb)}	%
\safemath{\dictsetgen}{\dictset_\text{gen}}
\safemath{\dictsetgenp}{\dictsetgen(\coh)}
\safemath{\dictsetonb}{\dictset_\text{onb}}
\safemath{\dictsetonbp}{\dictsetonb(\coh)}

\safemath{\leftside}{U}
\safemath{\rightsideA}{R_a}
\safemath{\rightsideB}{R_b}

\safemath{\indexS}{\setI_S} %

\safemath{\na}{n_a}			%
\safemath{\nb}{n_b}			%
\safemath{\coeffa}{p_i}	%
\safemath{\coeffb}{q_j}	%
\safemath{\seta}{\setP}		%
\safemath{\setb}{\setQ}     %
\safemath{\setw}{\setW}	%
\safemath{\setz}{\setZ}	%
\safemath{\cola}{\veca}		%
\safemath{\colb}{\vecb}		%
\safemath{\cold}{\vecd}		%
\safemath{\inputvec}{\vecx} 	%
\safemath{\error}{\vece}	%
\safemath{\noiseout}{\vecz} 	%
\safemath{\inputvecel}{x}
\safemath{\inputveca}{\vecx_a}
\safemath{\inputvecb}{\vecx_b}
\safemath{\outputvec}{\vecy}	%
\safemath{\lambdamin}{\lambda_{\mathrm{min}}}

\safemath{\elltwo}{\ell_2}
\safemath{\ellone}{\ell_1}
\safemath{\ellzero}{\ell_0}
\safemath{\ellinf}{\ell_\infty}
\safemath{\ellinftilde}{\ell_{\widetilde\infty}}
\safemath{\licard}{Z(\coh,\coha,\cohb)}
\safemath{\xsol}{\hat{x}}
\safemath{\xbord}{x_b}		%
\safemath{\xstat}{x_s}		%
\safemath{\xstatLone}{\tilde{x}_s}
\safemath{\order}{\mathcal{O}} %
\safemath{\scales}{\Theta} %
\safemath{\ones}{\mathbf{1}} %
\safemath{\zeroes}{\mathbf{0}} %
\safemath{\thlone}{\kappa(\coh,\cohb)} %
\safemath{\constoneA}{\delta} %
\safemath{\constoneB}{\epsilon} %
\safemath{\nlarge}{L}				   %
\safemath{\sumlarge}{S_\nlarge}
\safemath{\maxlarger}{P_\nlarge}	   %
\safemath{\Pzero}{\textrm{P0}}	
\safemath{\Pone}{\textrm{P1}}
\safemath{\vecfir}{\vecw}			 %
\safemath{\vecsec}{\vecz}
\safemath{\elvecfir}{w}              %
\safemath{\elvecsec}{z}				 %
\safemath{\nlargefir}{n}
\safemath{\normout}{\gamma}
\safemath{\auxfun}{h}
\safemath{\supp}{\textrm{supp}}%

\safemath{\indexa}{\ell}
\safemath{\indexb}{r}
\safemath{\indexc}{i}
\safemath{\indexd}{j}

\safemath{\project}{P}%

\newcommand{\fakeparagraph}[1]{\noindent {\bf #1 }}

\usepackage[capitalise,nameinlink]{cleveref}

\begin{document}

\title{Decentralized Multi-Agent Active Search and Tracking\\
when Targets Outnumber Agents %
}

\author{Arundhati Banerjee$^{1}$ and Jeff Schneider$^{1}$%
\thanks{$^{1}$ School of Computer Science,
        Carnegie Mellon University, Pittsburgh, USA. 
        {\tt\small arundhat@cs.cmu.edu}, {\tt\small schneide@cs.cmu.edu}}%
}

\maketitle

\begin{abstract}
Multi-agent multi-target tracking has a wide range of applications, including wildlife patrolling, security surveillance or environment monitoring. Such algorithms often make restrictive assumptions: the number of targets and/or their initial locations may be assumed known, or agents may be pre-assigned to monitor disjoint partitions of the environment, reducing the burden of exploration. This also limits applicability when there are fewer agents than targets, since agents are unable to continuously follow the targets in their fields of view. Multi-agent tracking algorithms additionally assume inter-agent synchronization of observations, or the presence of a central controller to coordinate joint actions. Instead, we focus on the setting of decentralized multi-agent, multi-target, simultaneous active search-\emph{and}-tracking with asynchronous inter-agent communication. Our proposed algorithm $\algoname$ uses a sequential monte carlo implementation of the probability hypothesis density filter for posterior inference combined with Thompson sampling for decentralized multi-agent decision making. We compare different action selection policies, focusing on scenarios where targets outnumber agents. In simulation, we demonstrate that $\algoname$ is robust to unreliable inter-agent communication and outperforms information-greedy baselines in terms of the Optimal Sub-Pattern Assignment (OSPA) metric for different numbers of targets and varying teamsizes.
\end{abstract}

\section{Introduction}
Searching for targets, detecting objects of interest (OOIs), localizing and following them are tasks integral to several robotics applications. 
Hence one or more of these sub-problems have been widely studied in a number of settings. 
For example, in informative path planning \cite{popovic2020informative} or simultaneous localization and mapping (SLAM) \cite{placed2023survey}, when the OOIs are fixed, an agent (or robot) adaptively selects actions to detect and localize the targets. 
Here the environment (or, the target distribution) being stationary, an agent tends to explore unseen parts of its surroundings more than exploit already observed viewpoints. 
In contrast, when targets are dynamic, the environment is non-stationary. 
Therefore, agents tracking an unknown number of moving targets should trade-off between exploring the possibly unobserved parts of the environment and exploiting its own posterior estimates to localize the previously detected targets at the current timestep. 
Unfortunately, prior work in multi-target tracking (MTT) has often assumed that the environment is known and exploration is not of primary concern \cite{robin2016multi}. 
Moreover, with multiple agents, existing MTT methods simplify the explore-exploit dilemma by separating search and tracking into sequential tasks where each agent is assigned to track a target as soon as it is found \cite{papaioannou2020cooperative}.  
Another approach is to assign sub-teams for executing these tasks separately \cite{chen2022active}. 
Further, the majority of these multi-agent multi-target tracking (MAMTT) algorithms require either a central controller to coordinate joint tracking actions, facilitate target hand-offs among agents or they depend on synchronized inter-agent communication for distributed inference and decision making.  
Unfortunately, such conditions may not be feasible in practice: the environment may be unstructured and unknown, OOIs may need to be simultaneously detected and localized (i.e. without a separation between search and tracking phases), there may not be enough agents to continuously monitor all the targets and unreliable communication channels may hinder inter-agent synchronization at each timestep.

In this work, we aim to develop a more practical approach to the MAMTT setup. 
In particular, we focus on the setting where agents are outnumbered by targets, so that the multi-agent team is unable to continuously cover all targets in their fields of view. %
The number of targets and their initial locations are unknown. 
Therefore, agents need to interact with the environment to collect observations by adaptively making explore-exploit decisions. %
It is not feasible to continuously track all targets, but our goal is to produce an estimate of the number and positions of all targets with time.
We assume that there is no central controller, and agents share their observations asynchronously with teammates, whenever possible. 
In other words, agents do not wait to receive observations, action selection policy or environment belief information from their teammates and can continue their online decision making when communication is unreliable or even unavailable.  

\fakeparagraph{Contributions.}
We propose a decentralized and asynchronous multi-agent algorithm, called $\algoname$ (\emph{Dec}entralized Multi-Agent Active \emph{S}earch-and-\emph{T}racking without continuous cov\emph{er}age) for simultaneous multi-target active search and tracking without continuously following the targets. 
In simulation, we compare a number of common decision making objectives from the tracking literature after adapting them to our simultaneous search-and-tracking setting. 
Our results show that $\algoname$ outperforms competitive baselines with different teamsizes and different target distributions in terms of the OSPA tracking performance.

\section{Related Works}
\label{sec:related_work}
Target detection and tracking are both widely studied problems,  
typically considered as distinct tasks in various applications like search and rescue \cite{murphy2004human}, security surveillance \cite{doitsidis2012optimal}, computer games \cite{oskam2009visibility}, etc. 
\cite{robin2016multi} is a detailed survey of the many different approaches and taxonomy used in robotics and related fields for such scenarios. 
Here, we discuss some of the most relevant prior work in our context. 

The single target state is commonly modeled assuming linear dynamics with additive Gaussian noise, using the Kalman filter \cite{kalman1960new} or using non-parametric particle filters \cite{doucet2001sequential}. 
In multi-target settings, alternative approaches like  Multiple Hypothesis Tracker (MHT) \cite{blackman2004multiple},  Joint Probabilistic Data Association (JPDA) \cite{fortmann1983sonar} and Probability Hypothesis Density (PHD) filter \cite{mahler2003multitarget} %
have been proposed, all of which differ in how they perform data association \cite{stone2013bayesian}. 
The PHD filter is particularly suited when unique identities for each target are not required, for example, in search and rescue tasks, where agents should detect and localize \emph{all} survivors. 
In this work, we build on the Sequential Monte Carlo (SMC) formulation of the PHD filter presented in \cite{ristic2010improved}.

Prior work in MAMTT algorithms primarily considers centralized or distributed settings, the latter still necessitating synchronized communication among subgroups of agents at each time step. 
Coupled with a PHD filter, some of the common action selection methods previously proposed for tracking include mutual information and expected count based objectives \cite{dames2017detecting}, Renyi divergence maximization \cite{papaioannou2020cooperative} and Lloyd's algorithm for Voronoi-cell based control \cite{dames2020distributed,chen2020distributed}.
\cite{papaioannou2020cooperative} discusses the benefits of a simultaneous search-and-tracking algorithm 
but their proposed method requires that agents transition from searching to tracking upon target detection, foregoing further exploration.
In similar spirit, \cite{van2022multi} proposes solving an information gain based multi-objective optimization problem over a finite planning horizon to decide the best (greedy) joint action for a centralized search-and-tracking task. 
In contrast with these deterministic objectives, \cite{xin2022comparing} demonstrates the superior performance of stochastic optimization methods like Particle Swarm Optimization (PSO) and Simulated Annealing (SA) for better coverage and localization in such settings. 
Unfortunately, none of these prior approaches are applied in the decentralized and asynchronous multi-agent setup when agents are unable to support continuous target coverage.

Recently, some learning based approaches have been proposed for tracking. 
\cite{jeong2021deep} considers the single-agent setting, assuming a known number of (one or two) targets which only start moving after first observed by the agent. 
\cite{zhou2022graph, tzes2023graph} both propose GNN-based algorithms, trained by imitation from a centralized expert, and deployed in the distributed inference and decision making with synchronized communication setup.  
While \cite{zhou2022graph} does not consider dynamic targets, \cite{tzes2023graph} simplifies the problem to deterministic optimal control over a fixed horizon, %
differing from our more complex setting.

\section{Problem Setup}
\label{sec:problem_setup}

\begin{figure}[htp]
  \centering
    \begin{subfigure}{0.5\linewidth}
    \centering
    \includegraphics[width=\textwidth]{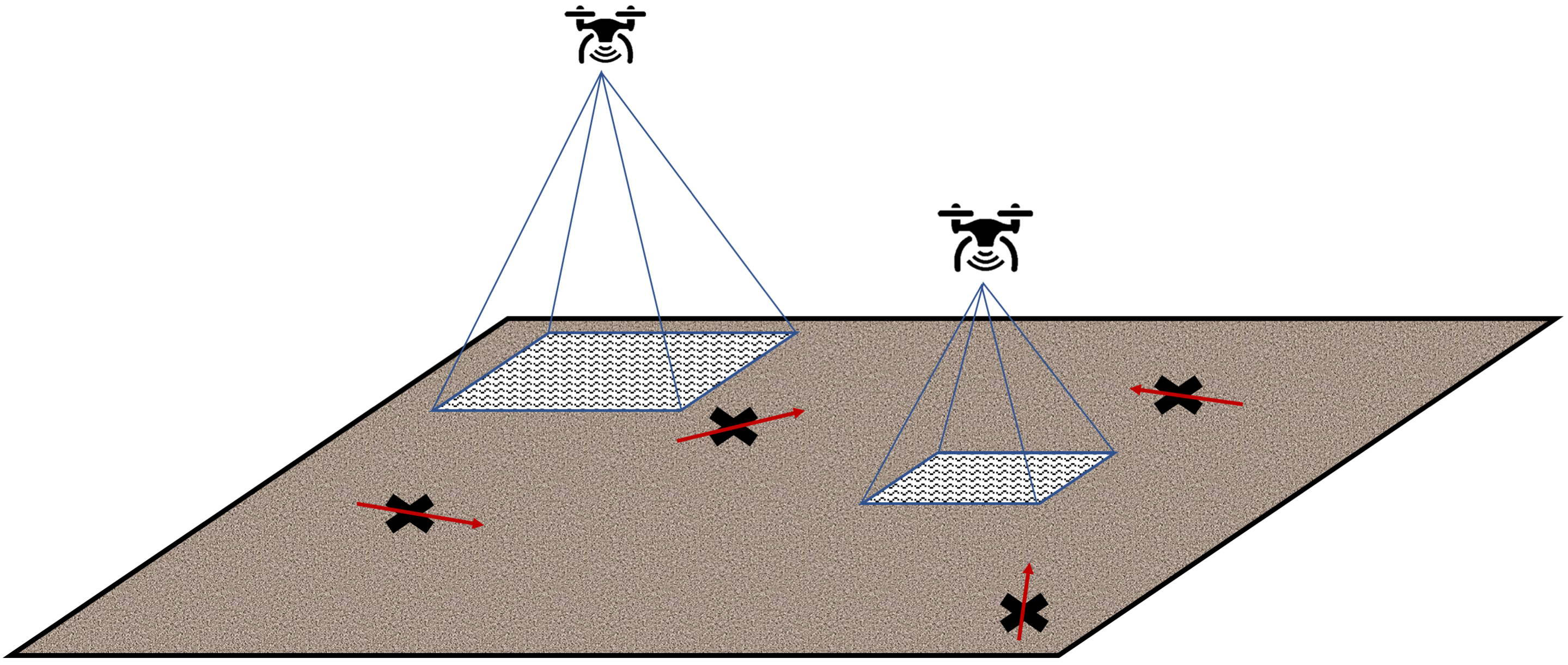}
    \caption{}
    \label{fig:searchspace}
    \end{subfigure}%
    \begin{subfigure}{0.5\linewidth}
    \centering
    \includegraphics[width=\textwidth]{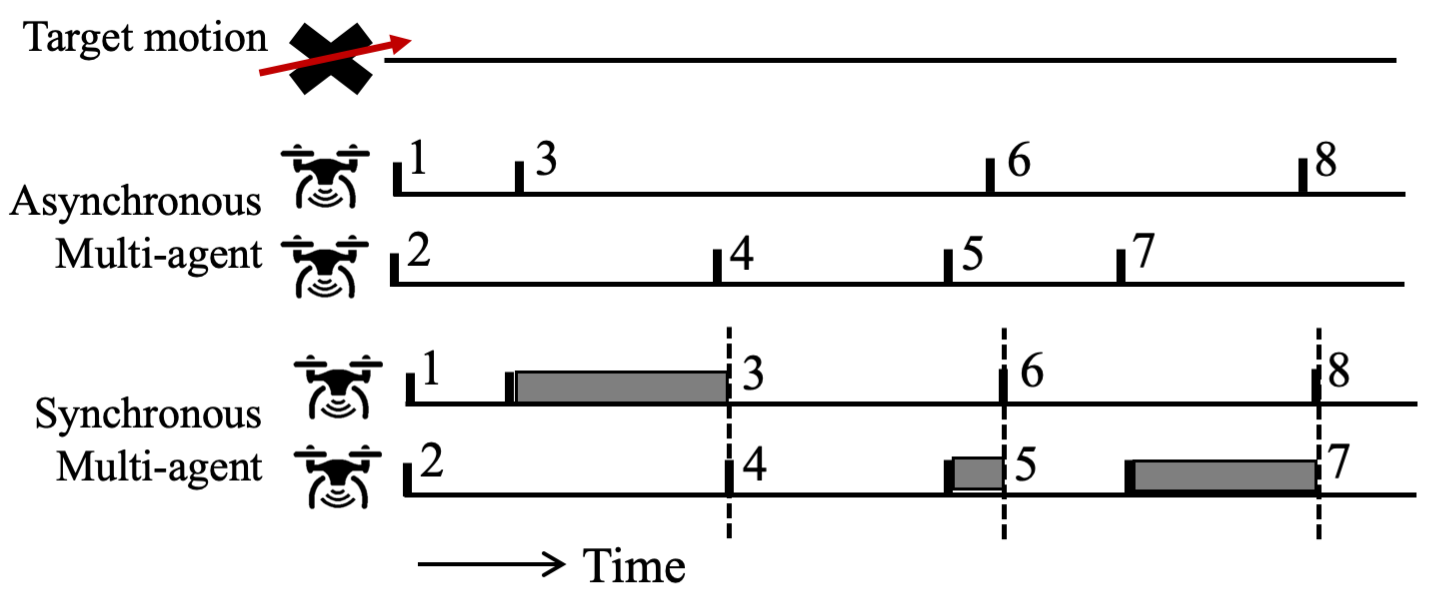}
    \caption{}
    \label{fig:async}
    \end{subfigure}
    \caption{\textbf{Problem setup.} (a) Agents sense different regions of the search space at different vertical heights, receiving noisy 2D location coordinates of the possible targets in their field of view, along with false positive measurements. The targets shown as black crosses move in the search space with different velocities shown by the red arrows. (b) The line at the top indicates the target's continuous motion with time. In our asynchronous multi-agent setup, agents can collect observations without waiting for their teammates whereas in the synchronous setting, the solid boxes indicate the agents' idle wait times.}
    \label{fig:setting}
\end{figure}

Consider a team of $\teamsize$ UAVs tasked with search and tracking of an unknown number of moving targets $\{1,\dots,\numtargets\}$ in a 2-dimensional (2D) region $\searchspace$ of length $\nlength$ and width $\nwidth$ (\cref{fig:searchspace}). 
The agents can self-localize. 
They are equipped with noisy sensors that provide location 2D coordinates of possible targets in their current field of view (FOV). 
The targets can move in any direction in the search space at different speeds. %
We assume that agents typically move faster than targets. %
Each agent's FOV includes a contiguous rectangular region of the search space, and agents may choose to observe a wider (smaller) area at a greater (lower) vertical height but with more (less) observation noise. 
We therefore consider a hierarchical region sensing action space for each agent. 
Agents can communicate %
asynchronously with their teammates. 
Over time $\totaltime$, agents observe different parts of the search space to detect and track all targets in the environment. 

\subsection{Target and Measurement Representations}
\label{subsec:target_dynamics}
The state of each target is denoted by $\targetstate = \begin{bmatrix}{\targetposx, \targetposy, \targetvelx, \targetvely}\end{bmatrix}^\transpose$, where 2D coordinates $(\targetposx,\targetposy) \in [0,\nwidth]\times[0,\nlength]$ and velocities $\targetvelx, \targetvely \in [-\maxspeed,\maxspeed]$. 
Since both the number of targets and their true locations are unknown, we follow the Random Finite Set (RFS) representation for the multi-target state space $\setX = \{\targetstate_1,\dots,\targetstate_{|\setX|}\}$, where $|\setX|$ follows a Poisson distribution and for a given cardinality, the set elements are sampled i.i.d from a uniform distribution \cite{mahler2014advances}. 
Following prior work, we use the Particle Hypothesis Density (PHD) filter \cite{mahler2003multitarget} to maintain a belief over the RFS \setX. 

The PHD $\nu(x)$ is the first statistical moment of a distribution over RFSs. 
In this case, it is a density over the state space of targets so that for any region $E \subseteq \searchspace$, the expected cardinality of the target RFS in that region is $\int_E\nu(x)dx$.  
The PHD filter tracks the evolving target density over the search space using models of target motion and measurements gathered by the agents. 
The measurements $\setZ$ are also modeled as a (Poisson) RFS, as are clutter $\kappa(z)$ (false positives) and target births $b(x)$. %

\subsection{Sensing model}
\label{subsec:sensing_model}
An agent with pose $\bmq = \begin{bmatrix}q_{x}, q_{y}\end{bmatrix}^\transpose$ executes a sensing action $\action_\bmq$, receiving a measurement set $\setZ = \{\bmz_1,\dots,\bmz_m\}$.  
Any target $\targetstate$ within the agent's FOV may generate a measurement $\bmz$, with a probability of detection $p_d(\targetstate|\bmq)$. 
In this work, we assume a constant $p_d(\cdot)$ when the target $\targetstate$ is within the FOV at $\bmq$, and $0$ otherwise.
The agent follows a linear sensing model with additive i.i.d white noise: $\bmz = \measurestate{\targetstate} + \bm\measurenoise$, where $\measurestate{\targetstate} = \begin{bmatrix}\targetposx, \targetposy\end{bmatrix}^\transpose$ and $\bm\measurenoise\sim\normal(\mathbf{0},\sigma_h^2\matidentity)$. 
Additionally, $\setZ$ includes i.i.d false positives with clutter rate $\lambda_\bmq$. %

The (noisy) target dynamics from state $\bm\xi$ to $\targetstate$ is captured by the target motion model $f(\bmx|\bm\xi)$. %
The survival probability $p_s(\targetstate)$ denotes the target's chances of persisting over successive time steps. 
The PHD filter formulates the following steps to propagate the posterior density over target states.\footnote{For a detailed understanding of the PHD filter, please refer to \cite{mahler2014advances}.} 
\begin{align}
    \text{Prediction: } &\bar{\nu}_{\itime}(\targetstate) = b(\targetstate) + \int_E f(\targetstate|\bm\xi)p_s(\bm\xi)\nu_{\itime-1}(\bm\xi)d\bm\xi\label{eq:phd_predict}\\
    \text{Update: } &\nu_\itime(\targetstate) = (1 - p_d(\targetstate|\bmq))\bar{\nu}_{\itime}(\targetstate) + \sum_{\bmz\in\setZ_t}\frac{\psi_{\bmz,\bmq}(\targetstate)\bar{\nu}_{\itime}(\targetstate)}{\eta_{\bmz}(\bar{\nu}_{\itime})}\label{eq:phd_update}\\
    &\eta_{\bmz}(\nu) = \kappa(\bmz|\bmq) + \int_E \psi_{\bmz,\bmq}(\targetstate)\nu(\targetstate)d\targetstate\\
    &\psi_{\bmz,\bmq}(\targetstate) = g(\bmz|\targetstate,\bmq)p_d(\targetstate|\bmq)
\end{align}
Here, $\psi_{\bmz,\bmq}(\targetstate)$ is the probability that the agent at $\bmq$ receives the measurement $\bmz$ from a target $\targetstate$ and $g(\bmz|\targetstate,\bmq)$ is the measurement likelihood model. 
The PHD filter can handle appearing and disappearing targets by defining an appropriate birth density $b(\targetstate)$ over the search space, but in our experiments the number of ground truth targets $k$ is fixed.   

\fakeparagraph{Remark 1.}
As \cite{mahler2003multitarget} explains, using the first order moment to approximate the multi-target belief and deriving recursive PHD update equations to approximate the evolving posterior is justifiable when both sensor covariances and false alarm densities are small, so that (the distribution of) observations from true targets are centered around target states with negligible spread and there is lower noise due to false alarms. 
Besides, in our SMC-PHD implementation following \cite{ristic2010improved}, we also ensure we avoid particle impoverishment %
during update and propagation of the posterior density estimate.

\fakeparagraph{Decentralized Asynchronous Multi-Agent Setup.} 
In our setup, each agent $\iagent$$\in$$\{1,\dots,\teamsize\}$ maintains its PHD estimate and decides its next sensing action in a decentralized manner. 
There is no central controller and inter-agent communication is asynchronous (\cref{fig:async}). 
Note that agents are not independent, since they share their own observations and incorporate the measurements received from their teammates in subsequent PHD filter update steps. %
Our setting is also different from the distributed computation setup in prior work \cite{dames2020distributed} %
where each agent completes a part of the centralized update step and relies on inter-agent synchronized communication to maintain a global PHD estimate across all agents. %

Since targets are in continuous motion, our agents must be able to deal with the uncertainty arising from observation noise as well as due to asynchronous communication of time-dependant observations in their posterior PHD updates. 
In order to enable time-ordered assimilation of received observations by all agents, we assume that any agent $\iagent$ communicates the tuple ($\itime,\action^{\bmq_\iagent}_\itime,\setZ^\iagent_\itime$) where $\action^{\bmq_\iagent}_\itime$ and $\setZ^\iagent_\itime$ are respectively the agent's sensing action and measurement set at time $\itime$.

\section{Our approach}
\label{sec:approach}

We now describe our algorithm $\algoname$ for multi-agent active search and tracking without continuous coverage. %

\fakeparagraph{Notation.}
Agent $\iagent$ at time $\itime$ has a history of available actions and observations $\dataset_{\itime}^\iagent = \{(\itime',\bma^{\bmq_\iagent'}_{\itime'}, \setZ^{\iagent'}_{\itime'})\}_{\itime'<\itime, \iagent'\in\{1,\dots,\teamsize\}}$.
Using $\dataset_{\itime}^\iagent$, it computes the PHD ${\nu}_\itime^\iagent$ (\cref{eq:phd_predict,eq:phd_update}) over the target RFS. 
In our SMC-PHD implementation, $\nu_\itime^\iagent = \{(w_{\itime,1}^\iagent,\targetstate_{\itime,1}^\iagent),\dots,(w_{\itime,\rho}^\iagent,\targetstate_{\itime,\rho}^\iagent)\}$ where $\targetstate_{\itime,1}^\iagent,\dots,\targetstate_{\itime,\rho}^\iagent$ are the $\rho$ particles with weights $w_{\itime,1}^\iagent,\dots,w_{\itime,\rho}^\iagent$. %

The SMC-PHD filter propagation steps follow from \cite{ristic2010improved}.\footnote{We will make our code publicly available after publication.} 
In our decentralized setup, each agent maintains its own posterior PHD $\nu_\itime^\iagent$. %
Next, we will describe the decision making step executed by agent $\iagent$ at time $\itime$. 

\fakeparagraph{Thompson sampling for decision making.} 
Prior work in multi-agent active search with static targets has demonstrated the effectiveness of Thompson sampling (TS) as a decentralized decision making algorithm, both in theory \cite{ghods2021decentralized} and in practice \cite{bakshi2023guts}. 
TS ensures stochasticity in decision making by sampling different plausible realizations of the ground truth from the posterior belief and selecting the best action to maximize the desired reward for a particular sample. 
The uncertainty in the agent's belief over the state space is reflected in the posterior samples, which makes TS suitable for driving exploration and exploitation. %
\cite{chen2022active} couples a TS-based active search strategy with the deterministic Lloyd's algorithm for tracking, but in their setup, agents are pre-assigned to only one of search or tracking tasks. %
Instead, we propose a TS strategy to enable agents to naturally trade-off exploratory sensing actions that might discover undetected targets, with exploitative sensing actions that help localize and track the previously detected dynamic targets in our %
simultaneous search-and-tracking setting.

To the best of our knowledge, prior work has not studied the problem of TS in a continuous (not discretized) search space with a PHD posterior. 
This is challenging because the PHD is not a distribution and does not include second order uncertainty information, whereas TS is typically applied in the Bayesian setting with the samples drawn from a posterior distribution for which both first and second order moment estimates are available \cite{russo2018tutorial}. 
Prior work in \cite{zhou2022regenerative} has proposed particle Thompson sampling (PTS) and regenerative PTS (RPTS) algorithms for particle filters where particles are sampled proportional to their weights. %
Therefore, we adopt a similar principle in our first proposed TS strategy for the SMC-PHD posterior, called TS-PHD-I (\cref{algo:TS1}).

\begin{algorithm}[htp]
    \caption{TS-PHD-I}
    \begin{algorithmic}[1]
    \State{\textbf{Input:} PHD $\nu = \{(w_1,\targetstate_1), \dots, (w_\rho,\targetstate_\rho)\}$.}%
    \State{Sample $\tilde{\rho}$ particles $\{\targetstate_i\}_{i=1}^{\tilde{\rho}}$ from $\nu$, proportional to the weights $\{w_1,\dots,w_\rho\}$}
    \State{Cluster the $\tilde{\rho}$ particles using k-means with $\tilde{n} = \sum_{i=1}^{\tilde{\rho}}w_i$ centroids ($\tilde{\setX} = \{\tilde{\targetstate}_1,\dots,\tilde{\targetstate}_{\tilde{n}}\}$) which form the TS}
    \end{algorithmic}
\label{algo:TS1}
\end{algorithm}

We note that this method has drawbacks. 
It tends to sample more particles from the %
regions in the PHD where the agent already estimates targets might be present. 
The samples drawn are thus more likely to be biased against 
regions of the target state space where the agent might be less certain about its observations owing to false positives or missed detections. 
Furthermore, this method does a poor job of modeling the uncertainty about the number of true targets.

To address the drawbacks of \cref{algo:TS1}, we now describe a second proposed approach to Thompson sampling from our SMC-PHD posterior (\cref{algo:TS2}). 
Recall that the expected cardinality of the target RFS $\setX$ over a region $E \subseteq \searchspace$ is given by $\hat{n} = \int_E \nu(\targetstate)d\targetstate$. 
In case of the SMC-PHD representation, $\hat{n} = \sum_i w_i, \forall \targetstate_i \in E$ \cite{ristic2010improved} i.e. the sum of particle weights of the SMC-PHD in the region $E$ is the expected cardinality of $\setX$ in that region. 
Further, \cite{mahler2003multitarget} shows that the PHD is the best Poisson approximation of the multitarget posterior in terms of KL divergence. 
We therefore draw a sample $\tilde{n}$ of the cardinality of the target RFS from a Poisson distribution with mean $\hat{n} = \sum_i w_i$ (i.e. $\tilde{n}\sim\text{Poisson}(\hat{n})$). %
Then we sample $\tilde{n}$ locations of the possible targets by drawing from a mixture of already estimated target locations in the PHD and some locations drawn uniformly at random over the search space. 
These $\tilde{n}$ particles $\tilde{\setX} = \{\tilde{\targetstate}_1,\dots,\tilde{\targetstate}_{\tilde{n}}\}$ form our TS.%

\begin{algorithm}[htp]
    \caption{TS-PHD-II}
    \begin{algorithmic}[1]
    \State{\textbf{Input:} PHD $\nu$$=$$\{(w_1,\targetstate_1), \dots, (w_\rho,\targetstate_\rho)\}$. $\hat{n}_\searchspace$$=$$\sum_{i=1}^\rho w_i$. Target location estimates $\hat{\setX}$$=$$\{\hat{\targetstate}_1,\dots,\hat{\targetstate}_{\hat{n}_\searchspace}\}$ from $\nu$.}
    \State{Sample $\tilde{n} \sim \text{Poisson}(\hat{n}_\searchspace)$}
    \State{Sample uniformly random target locations $\tilde{\setX}_R$}
    \State{Sample $\tilde{n}$ target locations ($\tilde{\setX}$) from $\hat{\setX}\cup\tilde{\setX}_R$ as the TS}
    \end{algorithmic}
\label{algo:TS2}
\end{algorithm}

\fakeparagraph{Objective.}
The Optimal Sub-Pattern Assignment (OSPA) metric is typically used in the MTT literature for evaluating the tracking performance of an algorithm and is defined as the error between two sets. 
Given sets $\setX$ and $\setY$, where $|\setX| = m \leq |\setY| = n$ without loss of generality, %
$$\text{OSPA}(\setX,\setY) = \big(\frac{1}{n}\min_{\pi\in\Pi_n} \sum_{i=1}^m d_c(x_i, y_{\pi(i)})^p + c^p(n - m)\big)^{\frac{1}{p}}$$ 
where $c$ is the cut-off distance, $d_c(x,y) = min(c, ||x-y||)$ and $\Pi_n$ is the set of all permutations of the set $\{1,\dots,n\}$. 
The distance error component of the OSPA computes the minimum cost assignment between $\setX$ and $\setY$, such that $x_i\in\setX$ is matched to $y_{i'}\in\setY$ only when they are within a distance $c$ of each other. 

Given a true target set $\setX = \{\targetstate_1,\dots,\targetstate_{|\setX|}\}$ and an estimated set $\setY = \{\bmy_1,\dots,\bmy_{|\setY|}\}$ of possible target locations, %
our goal is to minimize $\text{OSPA}(\setX,\setY)$. 
Since the ground truth $\setX$ is unknown, each agent $\iagent$ instead draws a TS $\tilde{\setX}_\itime^\iagent$ from the predicted PHD $\bar{\nu}_{\itime+1}^\iagent$. 
Assuming observations are generated by $\tilde{\setX}_\itime^\iagent$ for any action $\action$ and $\setY_\itime^\iagent$ is the estimated target set following the PHD filter update, agent $\iagent$ then selects:  
\begin{align}
\bma^\iagent_\itime = \argmin_\bma \expectation_{\setY_\itime^\iagent|\tilde{\setX}_\itime^\iagent,\bma} [\text{OSPA}(\tilde{\setX}_\itime^\iagent, \setY_\itime^\iagent)]\label{eq:objective}
\end{align} 
\cref{algo:our_algo} outlines our proposed algorithm, called $\algoname$. 
In our decentralized and asynchronous multi-agent %
setting, each agent individually runs $\algoname$ with its own sampled $\tilde{\setX}_\itime^\iagent$. %
Hence the stochasticity in the sampling procedure %
enables agents to make decentralized explore-exploit decisions for simultaneous search-and-tracking in their action space. 

\begin{algorithm}[htp]
    \caption{$\algoname$ for agent $\iagent$ at time $\itime$}
    \begin{algorithmic}[1]
    \State{\textbf{Input:} PHD $\nu_{\itime}^\iagent = \{(w_1^\iagent,\targetstate_1^\iagent),\dots,(w_\rho^\iagent,\targetstate_\rho^\iagent)\}$} 
    \State{Compute predicted PHD $\bar{\nu}_{\itime+1}^\iagent$ (\cref{eq:phd_predict}).}
    \State{Draw TS $\tilde{\setX}_\itime^\iagent\sim\bar{\nu}_{\itime+1}^\iagent$.}\label{line:TS}%
    \State{Assuming pseudo-measurements at $\tilde{\setX}_\itime^\iagent$, estimate expected target set $\setY_\itime^\iagent$ and select action $\action_\itime^\iagent$ (\cref{eq:objective}.}\label{line:decision}
    \State{Observe $\setZ_\itime^\iagent$. Update PHD $\nu_{\itime+1}^\iagent$ (\cref{eq:phd_update}).}
    \State{Estimate target set $\hat{\setX}_{\itime+1}^\iagent$ from $\nu_{\itime+1}^\iagent$ \cite{ristic2010improved}.}
    \State{Asynchronously communicate $(\itime, \action^\iagent_\itime, \setZ^\iagent_\itime)$ with team.}
    \end{algorithmic}
\label{algo:our_algo}
\end{algorithm}

\fakeparagraph{Remark 2.} 
Prior work in search-and-tracking \cite{papaioannou2020cooperative, chen2022active} tends to separate the search and tracking phases of the task, and maintains either a visit count or dynamic occupancy grid to compute the action selection objective during the exploration phase. %
Such methods scale poorly with the size of the environment since agents need to maintain a discretization over the search space \cite{van2022multi}. 
In contrast, our SMC inference for multi-target belief is parallelizable over particles in the posterior PHD, %
while our TS-based decision making is scalable with increasing teamsize $\teamsize$. %

\section{Results}
\label{sec:results}

We now describe our experimental setup. 
Consider a 2D search space with dimensions $n_l\times n_w = 16\times16$. 
There are $\numtargets$ targets moving in this region, whose starting locations and velocities are chosen uniformly at random, %
such that $v_{\max} = 0.1$. 
A team of $\teamsize$ agents are tasked with search-and-tracking of all the targets over $\totaltime = 150$ steps. 
The agents' action space $\setA$ consists of hierarchical region sensing actions of width $1\times1$, $2\times2$, $4\times4$ and $8\times8$, $|\setA| = 340$. 
Since actions with larger FOV receive noisier observations, we vary the false positive (clutter) rate as $\lambda \in \{0.005, 0.04, 1, 5\}$ for action widths 1, 2, 4 and 8 respectively. 
The survival probability in the PHD filter is set at $p_s = 1$ and the detection probability $p_d = 0.9$ for targets in the agent's FOV. 
In our SMC-PHD implementation following \cite{ristic2010improved}, we initialize 100 new (birth) particles per observation and re-sample 1000 particles per estimated target, following the low variance sampling method in \cite{thrun2002probabilistic}. 
We choose $\tilde{\rho}=100$ (\cref{algo:TS1}).

The agents assume the target motion model $\targetstate_{\itime+1} = \bF\targetstate_t + \bm\epsilon$, where $\bF = \begin{bmatrix}1&0&\Delta T&0\\0&1&0&\Delta T\\0&0&1&0\\0&0&0&1\end{bmatrix}$, $\Delta T = 1$ and $\bm\epsilon\sim\normal(\bm0,\bQ)$, $\bQ = \begin{bmatrix}
    0.03 & 0 & 0.05 & 0\\
    0 & 0.03 & 0 & 0.05\\
    0.05 & 0 & 0.1 & 0\\
    0 & 0.05 & 0 & 0.1
\end{bmatrix}$.
The sensing model is $\bmz = \bH\targetstate + \bm\omega$, where $\bH = \begin{bmatrix}
    1 & 0 & 0 & 0\\
    0 & 1 & 0 & 0
\end{bmatrix}$ 
and $\bm\omega \sim \normal(\bm0, \sigma^2\mathbf{I})$, $\sigma=0.1$. 
$\mathbf{I}$ is the identity matrix. 
For the OSPA metric, we set %
$c=2$ and $p=1$.

\fakeparagraph{Remark 3.} 
Our experimental setup is intended to %
illustrate the abilities of $\algoname$ for decentralized and asynchronous multi-agent multi-target search and tracking. %
The action space is chosen so that there is a non-trivial explore-exploit decision to be made by the agents. %
The maximum target speed is such that targets may cover a considerable distance in the search space over $\totaltime$ steps.

In the following experiments, we measure performance in terms of the average OSPA for the entire team of agents. %
The plots show mean across 10 random trials with the shaded regions indicating standard error. 
Each trial differs in the initialization of the target locations and their velocities. 

\subsection{Comparing TS-PHD-I with TS-PHD-II}
\label{subsec:compTS}

\begin{figure}[htp]
  \centering
    \begin{subfigure}{0.5\linewidth}
    \centering
    \includegraphics[width=\textwidth]{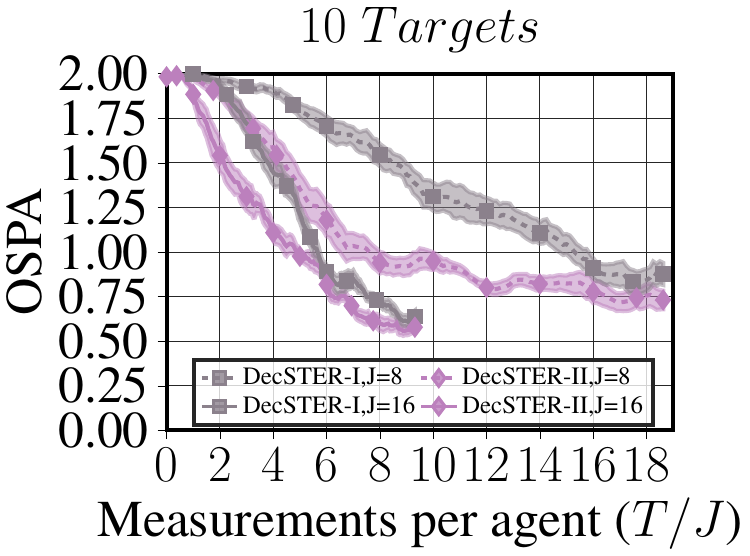}
    \label{fig:m1vsTbyJ}
    \end{subfigure}%
    \begin{subfigure}{0.5\linewidth}
    \centering
    \includegraphics[width=\textwidth]{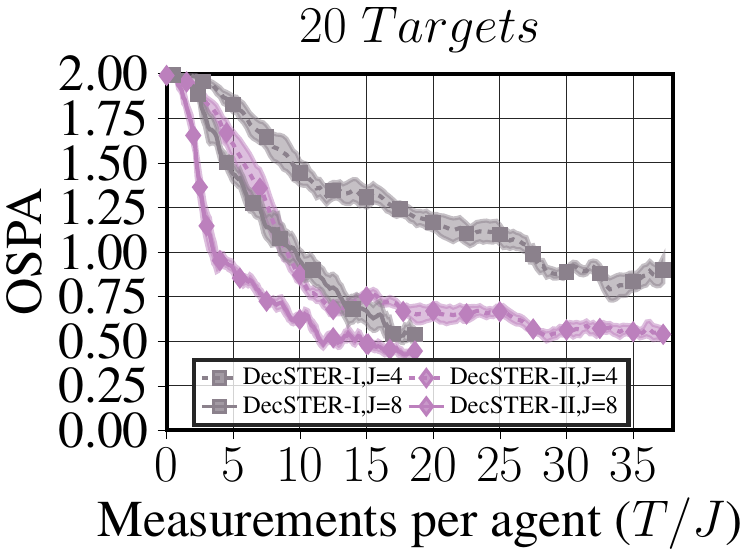}
    \label{fig:m2vsTbyJ}
    \end{subfigure}
    \caption{\textbf{Comparing the proposed TS methods}. $\algoname$-II using TS-PHD-II consistently outperforms $\algoname$-I using TS-PHD-I. Increasing team size $J$ reduces the number of measurements per agent required to achieve similar OSPA.}%
    \label{fig:results_m1vsm2}
\end{figure}

First, we compare the performance of $\algoname$ using the two proposed approaches for Thompson sampling from a PHD posterior. %
\cref{fig:results_m1vsm2} compares OSPA with number of measurements per agent, for $\algoname$-I and $\algoname$-II using TS-PHD-I and TS-PHD-II respectively in \cref{line:TS} of \cref{algo:our_algo}.  %
We observe that decision making with TS-PHD-II consistently outperforms that with TS-PHD-I. %
Since TS-PHD-II samples both the cardinality and locations of the target RFS from the PHD, the samples for different agents are sufficiently diverse to capture the uncertainty regarding the true multi-target ground truth. 
In contrast, the samples from TS-PHD-I are generally clustered around the agent's current estimate of target locations. 
We also empirically observed an improvement in the OSPA performance with TS-PHD-II when, for a particular sampled cardinality $\tilde{n}$, an agent considers multiple samples of $\tilde{n}$ target locations and averages the reward in \cref{eq:objective} over them. %
Our results using $\algoname$-II consider 10 such samples per action selection step for any agent $\iagent$. 
We also allow $\algoname$-I to similarly consider averaging over multiple samples, but this does not improve the sample diversity and does not lead to performance improvement.

\cref{fig:results_m1vsm2} further demonstrates the scalability of TS in the decentralized multi-agent active search-and-tracking setting.  
When teamsize increases $n$ times, agents achieve similar OSPA with $n$ times fewer measurements per agent \cite{ghods2021decentralized}.

\subsection{Baseline comparisons}
\label{subsec:baseline_comparison}
We compare $\algoname$-I and $\algoname$-II with the following baselines. 
Note that all of them use the same PHD filter inference method, but differ in the action selection policy. 

\fakeparagraph{1) RANDOM.} 
Each agent $\iagent$ selects its next sensing action uniformly at random. %
\fakeparagraph{2) RENYI.} 
At $\itime$, agent $\iagent$ computes the predicted PHD $\bar{\nu}_{\itime+1}^\iagent$ and generates a (pseudo) measurement set $\bar{\setZ}_\itime^\iagent$ for any action $\action\in\setA$ assuming the estimated target set $\hat{\setX}_\itime^\iagent$ from $\nu_\itime^\iagent$ as ground truth. 
It then selects the action $\action_\itime^\iagent$ that maximizes the Renyi divergence (with $\alpha=0.5$) between %
$\bar{\nu}_{\itime+1}^\iagent$ and its expected updated PHD %
${\nu'}_{\itime+1}^\iagent$ (\cref{eq:phd_update}). %
With the SMC-PHD formulation, the Renyi divergence is \cite{ristic2011note}: %
\begin{align}
\sum_{i=1}^\rho \bar{w}_i + \frac{\alpha}{1-\alpha}\sum_{i=1}^\rho {w'}_i - \frac{1}{1-\alpha}\sum_{i=1}^\rho {w'}_i^\alpha\bar{w}_i^{1-\alpha}\label{eq:renyi}
\end{align}
where $\bar{w}_i$ and ${w'}_i$ are the weights of the particle $i$ in $\bar{\nu}_{\itime+1}^\iagent$ and %
${\nu'}_{\itime+1}^\iagent$ respectively. 
\fakeparagraph{3) TS-RENYI.} %
We modify RENYI to use $\tilde{\setX}_\itime^\iagent\sim\bar{\nu}_{\itime+1}^\iagent$ (with TS-PHD-II) %
for computing the (pseudo) measurement set $\bar{\setZ}_\itime^\iagent$ and the updated weights ${w'}_i$. %

\begin{figure*}[htp]
  \centering
    \begin{subfigure}{0.25\linewidth}
    \centering
    \includegraphics[width=\textwidth]{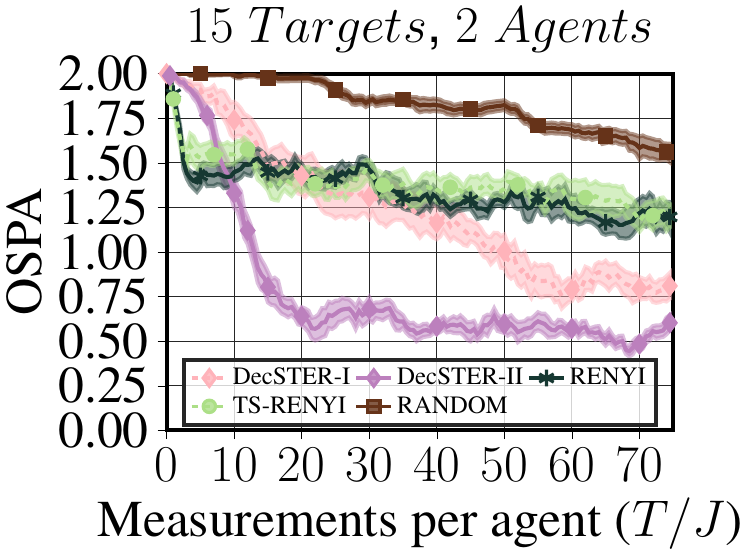}
    \label{fig:algos_k15_J2}
    \end{subfigure}%
    \begin{subfigure}{0.25\linewidth}
    \centering
    \includegraphics[width=\textwidth]{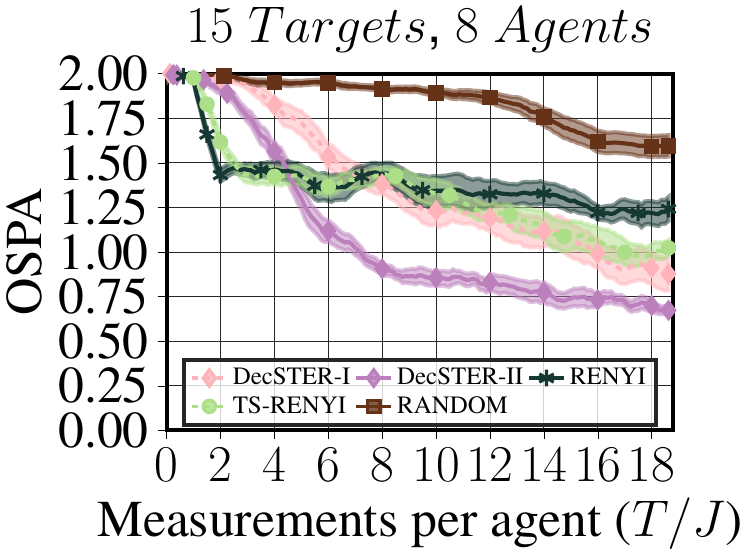}
    \label{fig:algos_k15_J8}
    \end{subfigure}%
    \begin{subfigure}{0.25\linewidth}
    \centering
    \includegraphics[width=\textwidth]{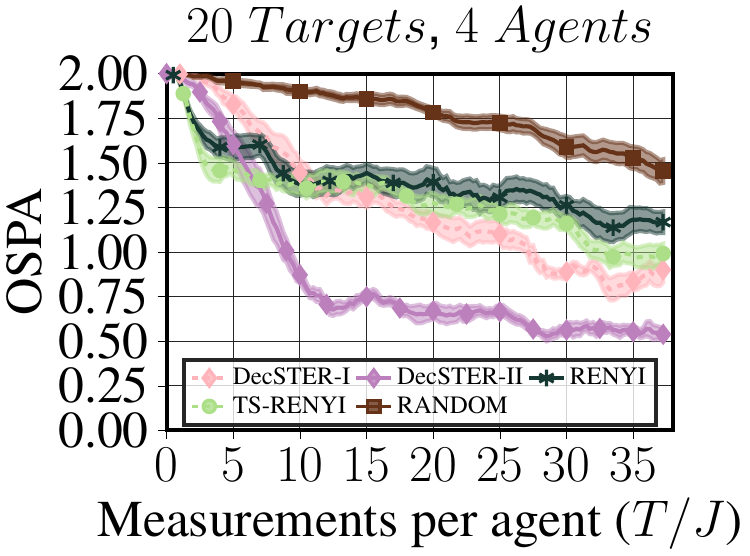}
    \label{fig:algos_k20_J4}
    \end{subfigure}%
    \begin{subfigure}{0.25\linewidth}
    \centering
    \includegraphics[width=\textwidth]{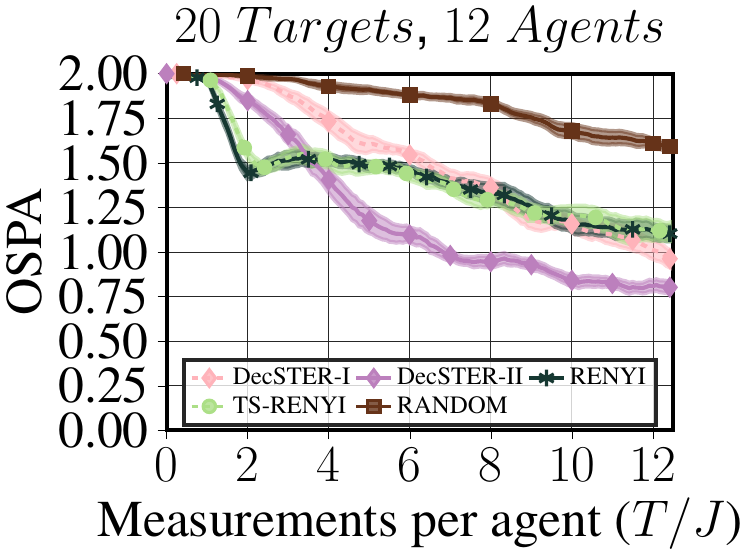}
    \label{fig:algos_k20_J12}
    \end{subfigure}
    \caption{\textbf{Baseline comparisons}. For different numbers of targets and with fewer agents than targets, $\algoname$ with TS-PHD-II (denoted $\algoname$-II) outperforms random sensing (RANDOM) and information greedy baselines (RENYI, TS-RENYI) by achieving a lower OSPA for the same number of measurements per agent. }%
    \label{fig:results_baselines}
\end{figure*}

\cref{fig:results_baselines} shows that our proposed method outperforms all the baselines for different number of targets $k$ and team sizes $J$. 
We observe that RENYI agents are information-greedy, therefore the lack of stochasticity in their decision making objective leads different agents to select the same action in the decentralized asynchronous multi-agent setting. 
Moreover, the computation in \cref{eq:renyi} depends only on the particles in $\bar{\nu}_{\itime+1}^\iagent$ and %
does not %
account for previously undetected targets. 
This highlights the drawback of using Renyi divergence as an optimization objective for explore-exploit decisions in the search-and-tracking setting, in contrast with its success in the tracking setting where exploration is not a concern \cite{papaioannou2020cooperative}. 
This motivated us to propose the TS-RENYI baseline in order to encourage exploration with samples drawn from TS-PHD-II. 
We observe that TS-RENYI still does not perform noticeably better than RENYI.
This is because the weights of the particles in the SMC-PHD filter relate to the expected cardinality of the target set, %
therefore \cref{eq:renyi} does not account for any measure of the distance error between ${\hat{\setX}}_\itime^\iagent$ (or ${\tilde{\setX}}_\itime^\iagent$) and the estimate ${\hat{\setX'}}_{\itime+1}^\iagent$ from ${\nu'}_{\itime+1}^\iagent$. %
In contrast, the OSPA objective accounts for both localization error as well as cardinality error in the estimated target set. %
Thus we observe that our algorithm $\algoname$-I is competitive with or outperforms random sensing and information-greedy baselines, and $\algoname$-II consistently achieves the lowest OSPA among all with the same number of measurements per agent. 
Based on these results, we consider $\algoname$-II as our best approach in this setting, labeled $\algoname$ in the following experiments.

\begin{figure}[htp]
  \centering
    \begin{subfigure}{0.5\linewidth}
    \centering
    \includegraphics[width=\textwidth]{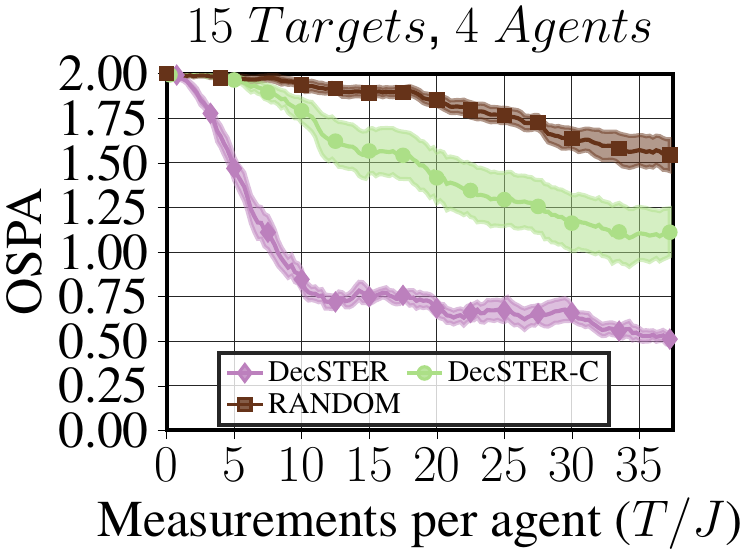}
    \label{fig:ablation_k15_J4}
    \end{subfigure}%
    \begin{subfigure}{0.5\linewidth}
    \centering
    \includegraphics[width=\textwidth]{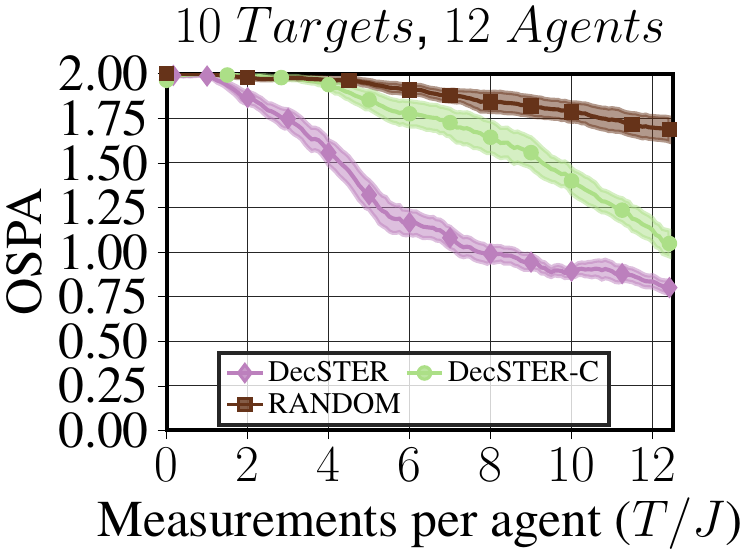}
    \label{fig:ablation_k20_J16}
    \end{subfigure}
    \caption{\textbf{DecSTER vs. DecSTER-C}. $\algoname$-C optimizes only for the cardinality error in the OSPA objective. $\algoname$ outperforms $\algoname$-C indicating the advantage of jointly optimizing detection and localization errors with TS-guided explore-exploit decisions.}
    \label{fig:results_ablations}
\end{figure}

\fakeparagraph{DecSTER vs. DecSTER-C.} 
Motivated by prior work \cite{dames2017detecting} that showed the effectiveness of maximizing the expected number of target detections for action selection in multi-agent tracking, we introduce the $\algoname$-C baseline where agents select actions minimizing only the cardinality error term of the OSPA. 
Unlike $\algoname$, each agent with $\algoname$-C draws multiple samples of cardinality $\tilde{n}\sim \text{Poisson}(\hat{n})$ to consider the uncertainty about the real number of targets in its objective. 
\cref{fig:results_ablations} shows that $\algoname$ still outperforms $\algoname$-C, indicating that jointly considering detection and localization error in the decision making objective is more advantageous for TS-guided action selection in our search-and-tracking setting. 
This further supports our earlier observations regarding the drawbacks of using the particle weight based Renyi divergence optimization objective in this setting.

\subsection{Robustness to communication delays}
\label{subsec:exp_comm}

\begin{figure}[htp]
  \centering
    \begin{subfigure}{0.5\linewidth}
    \centering
    \includegraphics[width=\textwidth]{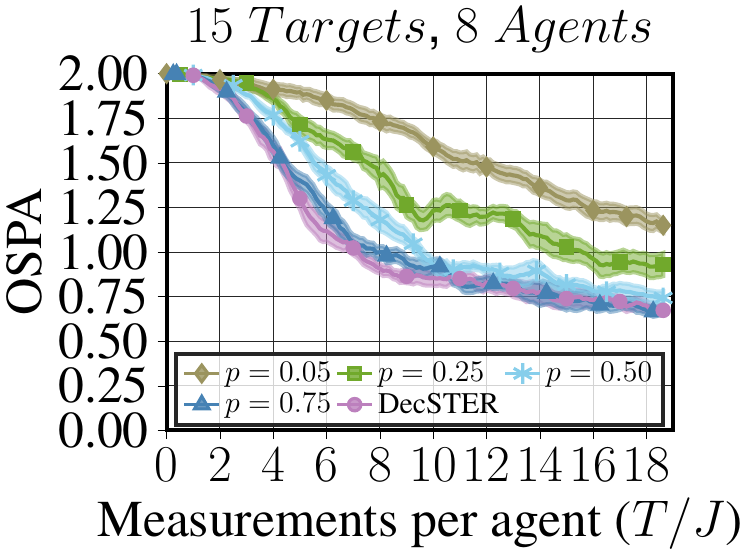}
    \label{fig:comm_k15_J8}
    \end{subfigure}%
    \begin{subfigure}{0.5\linewidth}
    \centering
    \includegraphics[width=\textwidth]{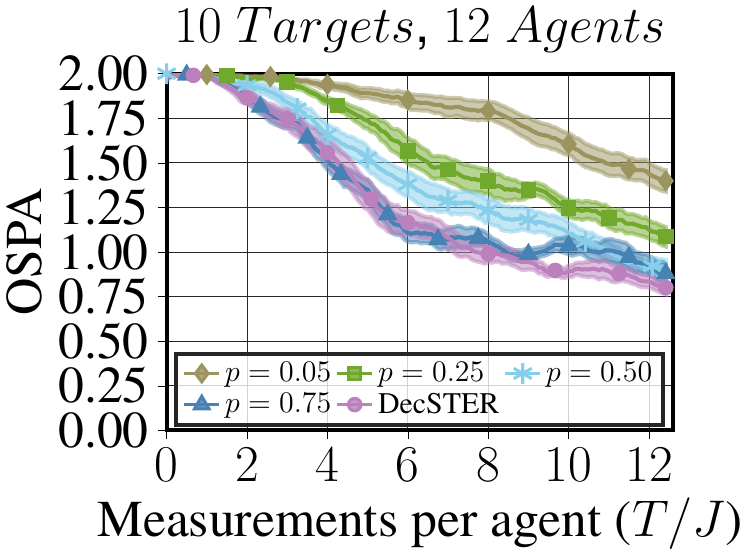}
    \label{fig:comm_k10_J12}
    \end{subfigure}
    \caption{\textbf{Robustness to unreliable communication}. When agents communicate their actions and observations with decreasing probability $p$, $\algoname$ experiences a graceful deterioration in OSPA performance and agents require increasingly more measurements to estimate the number and locations of true targets in the search space.}%
    \label{fig:results_comm_delay}
\end{figure}

Multi-agent systems benefit from leveraging observations shared by their teammates. 
Agents in our decentralized and asynchronous multi-agent setting benefit from any information shared by their teammates but they can continue searching for and tracking targets without waiting for such communication. %
Therefore, we now analyze the robustness of $\algoname$ under unreliable inter-agent communication. 
In simulation, we consider each agent chooses to communicate its own observation at time $\itime$, along with any prior observations it had not shared with its teammates, with a probability $p \in \{0.05, 0.25, 0.50, 0.75, 1\}$. 
The $p = 1$ setting corresponds to our description and analysis of $\algoname$ in \cref{fig:results_baselines}. 
We observe a graceful decay in the OSPA performance with decreasing rates of inter-agent communication in \cref{fig:results_comm_delay}, both when targets outnumber agents and vice versa.  
Compared to prior work in the centralized or distributed multi-agent tracking setting \cite{robin2016multi}, $\algoname$ does not depend on synchronized communication within the team, thus agents can adapt and continue their search-and-tracking tasks even when communication is unreliable or unavailable. %

\section{Conclusion}

We introduce $\algoname$, a novel decentralized and asynchronous algorithm for multi-agent multi-target active search-and-tracking that relaxes the restrictive assumption of requiring continuous target coverage. 
In simulation, $\algoname$ outperforms competitive baselines that greedily optimize for information gain or expected target detections. 
A key contribution is adapting TS to effectively drive exploration and exploitation using the SMC-PHD filter. %
Future work includes theoretical analysis of the proposed TS methods and learning improved models of environment uncertainty for non-stationary multi-target tracking. 
Building on recent success of TS-based multi-agent active search \cite{bakshi2023guts}, we also aim to validate $\algoname$'s performance when deployed on teams of physical robots in the real world.

\bibliographystyle{IEEEtran}
\bibliography{IEEEEabbrv,ref}

\end{document}